\documentclass[preprint,12pt]{elsarticle}

\usepackage{amsmath} 
\usepackage{algorithmic}

 \usepackage{algorithm}

\newcommand{\fracpartial}[2]{\frac{\partial {#1}}{\partial {#2}}}
\newcommand{\fpevalat}[3]{\left(\fracpartial{#1}{#2}\right)_{#3}}
\newcommand{\bigDevalat}[3]{\left(\bigD{#1}{#2}\right)_{#3}}
\newcommand{\bigD}[2]{\frac{D {#1} }{ D {#2}}}
\newcommand{\vecDelta}{\vec{\delta}}
\newcommand{\df}[0]{} 
\newcommand{\dfa}[1]{}
\newcommand{\Gtarget}{{G'}}
\newcommand{\Vtarget}{{V'}}
\newcommand{\Vapprox}{{\widetilde{V}}}
\newcommand{\Gapprox}{{\widetilde{G}}}
\newcommand{\Qapprox}{{\widetilde{Q}}}
\newcommand{\vecx}{\vec{x}}
\newcommand{\veca}{\vec{a}}
\newcommand{\dimWeights}{n_w}
\newcommand{\dimX}{n_x}

\newcommand{\ActionSpace}{A}
\newcommand{\Weights}{\vec{w}}
\newcommand{\Weightsz}{\vec{z}}

\newcommand{\Vpi}{V^{\pi}}
\newcommand{\Rpi}{V^{\pi}}
\newcommand{\RpiOpt}{V^{*}}
\newcommand{\ua}{u(\veca)}
\newcommand{\ddQduaSqaredt}{\left(\frac{\partial {^2\Qapprox}}{\partial \ua \partial \ua}\right)_t}
\newcommand{\ddQdaSqaredt}{\left(\frac{\partial {^2\Qapprox}}{\partial \veca\partial \veca}\right)_t}
\newcommand{\BlackBox}{\rule{1.5ex}{1.5ex}}  

\newtheorem{thm}{Theorem}
\newproof{pf}{Proof}
\newtheorem{cor}{Corollary}
\newdefinition{rmk}{Remark}
\newtheorem{lem}{Lemma}

\journal{Arxiv}

\begin{document}

\begin{frontmatter}



\title{The Local Optimality of Reinforcement Learning by Value Gradients, and its Relationship to Policy Gradient Learning}


\author{Michael Fairbank}
\ead{michael.fairbank.1 `at' city.ac.uk}

\author{Eduardo Alonso}
\ead{eduardo `at' soi.city.ac.uk}
\address{Department of Computing, School of Informatics, City University London, London, UK}

\begin{abstract}
In this theoretical paper we are concerned with the problem of learning a value function by a smooth general function approximator, to solve a deterministic episodic control problem in a large continuous state space.  It is shown that learning the gradient of the value-function at every point along a trajectory generated by a greedy policy is a sufficient condition for the trajectory to be locally extremal, and often locally optimal, and we argue that this brings greater efficiency to value-function learning.  This contrasts to traditional value-function learning in which the value-function must be learnt over the whole of state space.  

It is also proven that policy-gradient learning applied to a greedy policy on a value-function produces a weight update equivalent to a value-gradient weight update, which provides a surprising connection between these two alternative paradigms of reinforcement learning, and a convergence proof for control problems with a value function represented by a general smooth function approximator.

\end{abstract}


\end{frontmatter}


\section{Introduction}

Reinforcement learning (RL) is the study of how an agent can learn actions that maximise some given reward function. For example a typical scenario is a robot (or agent) wandering around in an environment, such that at time $t$ it has position (or state) vector $\vecx_t$.  The robot moves in continuous space and time, but we discretize time to enable modelling of the motion by a computer.  Thus at each time $t$ the robot chooses an action $\veca_t$ which takes it to the next state according to the environment's model function $\vecx_{t+1}=f(\vecx_t, \veca_t)$, and gives it an immediate scalar reward $r_t$, given by the reward function $r_t=r(\vecx_t, \veca_t)$.  The robot keeps moving until it reaches one of the designated terminal states.  The RL problem is for the robot to learn how to choose actions so as to maximise the total reward received, $\Sigma_t r_t$.  Specifically, the problem is to find a {\it policy} function $\pi(\vecx, \Weightsz)$ (where $\Weightsz$ is some parameter vector) that calculates which action $\veca=\pi(\vecx,\Weightsz)$ to take for any given state $\vecx$, such that the total reward is maximised.

One key approach to tackle this RL problem is to assign a score to every point in state space that gives the best possible {\em total} reward attainable if starting from that state.  This scoring function is called the {\em optimal value function}, $\RpiOpt(\vecx)$.  If this function was perfectly known then it would be easy for the robot to behave optimally because at any instant it could consider all possible actions available and always choose the one that leads to the best valued state, whilst also taking into account the immediate short-term reward in getting there.  This way of acting is called the {\em greedy policy} on $\RpiOpt$.  So the objective of learning is to make a function approximator, $\Vapprox(\vecx, \Weights)$ (e.g. a neural network with weight vector $\Weights$), learn and represent the optimal value function, and then use a greedy policy on the approximated function.

However the optimal value function is not known at the start of learning.  So for {\em any} given policy $\pi(\vecx, \Weightsz)$ we can define its value function $\Vpi(\vecx, \Weightsz)$ to be the real valued total reward that would be encountered if the robot started at state $\vecx$ and followed that policy until termination.  Bellman's Optimality Condition \citep{bellman57} shows that if $\Vapprox\equiv\Vpi$ for all $\vecx$ in the state space $S$,  where $\pi$ is the greedy policy on $\Vapprox$, then that greedy policy is optimal.   There is a circular interdependence here; $\Vpi$ depends on the greedy policy $\pi$, which depends on $\Vapprox$, and we want $\Vapprox\equiv\Vpi$ for all $\vecx$.

If the state space was discrete and finite then Bellman's condition could be met by dynamic programming which makes iterative sweeps through the whole of state space, updating $\Vapprox$ incrementally.  But in our problem the state space is large and continuous so this is not possible.  The RL methods TD(0) and Q-learning \citep{suttonbarto-1998, watkins92} can be used to update $\Vapprox$ along one trajectory at a time, but these can be very slow since Bellman's condition still needs meeting over the entire state space for optimality.  Even if Bellman's condition is perfectly satisfied along a single trajectory, performance can be extremely far from optimal if Bellman's condition is not satisfied over the neighbouring trajectories too.   Hence it is well known in the RL community that constant exploration of the environment must be applied.  This exploration could be provided by stochastic model functions, a stochastic policy, or a stochastic start point for each trajectory.  The ability of RL algorithms to work in stochastic environments is a virtue, but it is also a necessity for the above reason, and it is a goal of this paper to define value-function learning methods that work in a deterministic environment.

TD($\lambda$) \cite{sutton88learning} is a generalization to TD(0) which uses an extra parameter $\lambda\in[0,1]$ that can improve the speed of learning.  The effect of $\lambda$ is described in detail in section \ref{sec:targetValues}, where we call it the ``bootstrapping'' parameter.  

Although value function learning methods have produced successes in robot control \citep{Kwok04reinforcementlearning, doya00reinforcement}, value function learning methods are problematic in that their theoretical convergence guarantees with function approximators are limited.  TD($\lambda$) has been proven to converge \cite{tsitsiklis96analysis} provided the function approximator for $\Vapprox(\vecx, \Weights)$ is {\em linear in $\Weights$}, and the policy is fixed (i.e. that excludes the greedy policy on $\Vapprox$).   They are {\it not proven} to converge when a general function approximator is used to represent the value function (e.g. a neural network) or when a greedy policy is used, such as is required by our robot RL problem.    Divergence examples exist for a non-linear function approximator \cite{ tsitsiklis96analysis}, and where $\Vapprox$ is linear in $\Weights$ but where a greedy policy is used (diverging for both $\lambda=0$ and $\lambda=1$; see section 4.3 of \cite{fairbank08}).

One reason that these methods do not always converge is that changing the approximated value function $\Vapprox$ at one point in state space will cause $\Vapprox$ to change in other points of state space too, since the function approximator that represents $\Vapprox(\vecx, \Weights)$ cannot be infinitely flexible.  A second reason is that in the Bellman condition, $\Vpi$ depends on $\pi$ which in turn depends on $\Vpi$, so making progress in learning one of them can undo progress in learning the other.  This second issue is highly relevant for RL control problems, since the ultimate objective is not just to learn a value function for some {\it fixed} policy, but is to improve a policy until it becomes optimal (or close enough to optimal).  Thus any successful convergence analysis for value function learning must cope with the concurrent updating of $\Vapprox$ and the greedy policy, and there have been few insights into this problem by the RL literature--- convergence proofs so far have generally treated one of these two components as fixed, or only treated the tabular case.  

We address these issues by following the method of Dual Heuristic Programming (DHP) of Werbos \citep{hic92} which tries to explicitly learn the gradient of the value function with respect to the state vector, i.e. it learns $\fracpartial{\Vpi}{\vecx}$ instead of $\Vpi$ directly.  We call this method {\it value gradient learning} (VGL), to distinguish it from the usual direct updates to the values of the value function, which we refer to as {\it value learning} methods (VL).  We extend Werbos' method to include a bootstrapping parameter $\lambda$ (just as Sutton did in extending TD(0) into TD($\lambda$)), to give the algorithm we call VGL($\lambda$).

The VGL method addresses the issue of the Bellman equation needing to be solved over the whole of state space, in that it turns out to be only necessary to learn the value gradient along a {\it single trajectory} for it to be locally optimal.  This contrasts strongly with the VL methods which need to learn the value function over all immediately neighbouring trajectories too for local optimality, and so this is a significant efficiency gain for the VGL method.  This optimality is an almost-immediate consequence of Pontryagin's maximum principle  \citep{pontryagin}, and this is proven in Section \ref{sec:optimality}.    

We address the difficulty of analysing the interdependence of simultaneously updating $\Vpi$ and $\pi$ by showing (in  Lemma \ref{lem:lemma2}) that the dependency of a greedy policy on a value function is {\it primarily through the value-gradient}.  Hence a value-gradient analysis is necessary at some level to provide a theoretical gateway to analysing the convergence properties of any value-function weight update that uses a greedy policy; be it a VGL weight update or a VL weight update. 

The dependency of the greedy policy on the value-gradient has already been exploited in an efficient policy \cite{doya00reinforcement}, but the VGL method takes this one step further by trying to explicitly learn the value-gradient.

There is an alternative paradigm of RL called policy gradient learning (PGL) which does not rely on learning a value function at all.  We define PGL as algorithms that do gradient ascent on the total reward, and this definition includes methods of \cite{williams, backproptime90, Peters06policygradient, munos06}.  These methods have natural convergence guarantees since they are hill climbing strategies on a function with an upper bound, and have proved successful at robot control in continuous spaces \citep{Peters06policygradient}. 

In Section \ref{sec:pgl}, we show a VGL weight update with $\lambda=1$ is identical to a PGL weight update, and this makes a theoretical connection between these two different paradigms of RL, and provides a convergence proof for this value function control problem with a {\it general} function approximator (provided $\lambda=1$ and provided the policy is differentiable at every time step of the trajectory).

In summary, the VGL methods in this paper lead to several benefits and insights: 
\begin{itemize}
 \item They make a more direct approach to RL since it is the value-gradient that affects the policy 
and so it makes sense that this is what should be learned.
 \item It is only necessary to learn the value-gradient along a single trajectory instead of the whole of state space.  This can lead to improved efficiency for VGL methods, since there is no need to explore locally (see Section \ref{sec:optimalityDiscussion}).
 \item They provide a theoretical insight and convergence result into the long-standing problem of proving convergence for value-function learning methods with a function approximator, while providing a theoretical link between value-function learning and PGL.
\end{itemize}

Another goal of this paper is to raise awareness in the RL community of the methods of \cite{hic92}.

The VGL method is a ``model based'' RL method in that it requires that the model functions $f(\vecx, \veca)$ and $r(\vecx, \veca)$ to be known.  Knowledge of the model functions (and their derivatives) delivers many of the above benefits of the VGL method.  Many researchers define RL specifically for the case of unknown model functions.  To answer this we would have to supplement this VGL method with a separate learning system specifically to learn the model functions, prior to trying to learn the policy.  This is a commonly used strategy for successful RL methods, for example in the recent success of maintaining the inverted flight of a helicopter \cite{ng04}, the model functions were learned separately and prior to learning a policy.  We also suggest that model learning is the {\it relatively} straightforward part of the RL task, since it is a supervised learning problem, where the immediate answers are given. Also the model functions in a control problem are often simple known laws of physics, and they do not change much from point to point, due to the continuous nature of the environment.  \cite{munos06} exploits this to successfully learn the model functions, in real time, entirely while the robot is travelling along one single trajectory.  However we acknowledge that ideally, for a full RL learning system, we would concurrently learn the model, the policy, and the value function while still ensuring convergence.  In this paper we successfully analyse items two and three of this list in the case of $\lambda=1$, but at the expense of assuming the first is fixed and known.

We note that a third paradigm of RL exists called an {\em actor-critic} architecture.  In this architecture there is one function approximator to represent $\Vapprox$, and a second function approximator to represent the policy.  The greedy policy is not used.  Successful theoretical results exist for the concurrent updating of the policy and $\Vapprox$ \citep{suttonMcallester}, and we discuss these results in comparison to our own in section \ref{sec:pglEquivalenceDiscussion}.

This paper is organised as follows.  Section \ref{sec:vglForRL} defines the VGL($\lambda$) algorithm and gives the definitions necessary to do this.  Key concepts defined there are the approximated value function and its target values; and the approximated value gradient and its target.  The next two sections give the two main theoretical results:  In Section \ref{sec:optimality} we prove the local optimality of the objective of VGL for a single trajectory and discuss the efficiency of the method.  In Section \ref{sec:pgl} we demonstrate the connection between VGL and PGL, and hence give the convergence proof for VGL with $\lambda=1$, under certain conditions.  Short discussions follow each of the two main theoretical results.  Section \ref{sec:conclusions} presents the conclusions of our work.

\section{Value Gradient Learning for Reinforcement Learning} \label{sec:vglForRL}
This section defines the VGL algorithm.  After some preliminary definitions are made in section \ref{sec:definitions}, we describe target values which can be used to define VL (in section \ref{sec:targetValues}).  This definition of target values and VL is done a concise way that differs from the conventional RL literature, and it allows us to define the VGL targets (in section \ref{sec:targetValueGradient}) and the VGL algorithm (section \ref{sec:theVGLalgorithm}).  Both of these VGL concepts will be new to readers only experienced with VL.  A technical difficulty that needs dealing with on the way are saturated actions, which are defined in section \ref{sec:saturatedActions}.

\subsection{Preliminary Definitions}
\label{sec:definitions}
{\bf \noindent State space, trajectories and model functions.}
{\it State Space}, $S$, is a subset of $\Re^n$. 
Each state in the state space is denoted by a column vector $\vecx$.  The state space is large and continuous so that a function approximator is necessary to represent the learned policy.  A {\it trajectory} is a list of states $(\vecx_0, \vecx_1, \ldots, \vecx_F)$ through state space starting at a given point $\vecx_0$.  The trajectory is parameterised by actions ${\veca_t}$, chosen from an action space $\ActionSpace$, for time steps $t$ according to a model.   The {\it model} is comprised of two known smooth deterministic functions $f(\vecx, \veca)$ and $r(\vecx, \veca)$.  The first model function $f$ links one state in the trajectory to the next, given action $\veca_t$, via the ``Markovian'' rule $\vecx_{t+1}=f(\vecx_t, \veca_t)$.  The second model function, $r$, gives an immediate real-valued reward $r_t=r(\vecx_t, \veca_t)$ on arriving at the next state $\vecx_{t+1}$.  

Some states in $S$ are designated as terminal states.  Assume that each trajectory is guaranteed to reach a terminal state in some finite time (i.e. the problem is episodic).   For example, a scenario like this could be an aircraft with limited fuel trying to land; or it could be a navigation problem with an imposed time limit.  For a particular trajectory label the final time step $t=F$, so that $\vecx_F$ is the terminal state of that trajectory.  Note that for a general trajectory, $F$ is dependent on the start point and the actions taken, so is not a global constant.

{\bf \noindent Action vectors.}
The action vectors $\veca$ can be real scalars or have several real components, one for each of the control dimensions of the agent.  For example in a car, the action components might be accelerator pedal position, steering wheel angle, and brake pedal position.  For a monorail train, there might be just one scalar needed.  Assume each action component $(\veca_t)^i$ is a real number that, for some problems, may be constrained to $(\veca_t)^i \in [-1,1]$, and these constraints are imposed by the action space, such that $\veca_t \in \ActionSpace$ for all $t$.\footnote{The choice of the range $[-1,1]$ is arbitrary.  The theoretical results of this paper would also apply to any other finite range.}

{\noindent \bf Policy.}  A policy is a function $\pi(\vecx, \Weightsz)$, parameterised by a vector $\Weightsz$, that generates actions as a function of state.  Thus for a given trajectory generated by a given policy $\pi$, $\veca_t=\pi(\vecx_t, \Weightsz)$. Since the policy is a pure function of $\vecx$ and $\Weightsz$, the policy is memoryless.  The vector $\Weightsz$ holds the parameters of a smooth function approximator, for example it could be a concatenation in column vector form of all of the weights of a neural network. 

{\noindent \bf Value Function.} If a trajectory starts at state $\vecx_0$ and then follows a policy  $\pi(\vecx, \Weightsz)$ until reaching a terminal state, then the value function $\Rpi(\vecx_0, \Weightsz)$ returns the total reward received:
\begin{eqnarray}
\Rpi(\vecx_0, \Weightsz) &=&\sum_{t = 0}^{F-1} \dfa{^t} r(\vecx _{t}, \pi(\vecx_t,\Weightsz)) \nonumber\\
&=&r(\vecx_0, \pi(\vecx_0,\Weightsz))+\df \Rpi(f(\vecx_0, \pi(\vecx_0,\Weightsz)), \Weightsz) \label{eqn:Rpi_definition}\end{eqnarray}
with $\Rpi(\vecx_F,\Weightsz)=0$.

{\noindent \bf Approximate Value Function.}  We define $\Vapprox(\vecx,\Weights)$ to be the real-valued output of a smooth function approximator with weight vector $\Weights$ and input vector $\vecx$.  This is the approximate value function.  
It is the intention of most VL algorithms to eventually find $\Weights$ such that $\Vapprox(\vecx,\Weights) \approx \RpiOpt(\vecx)$ for all $\vecx$, as described earlier.

{\noindent \bf Approximate Value-Gradient.}  The approximate value-gradient function $\Gapprox(\vecx,\Weights)$ is defined to be $\Gapprox(\vecx,\Weights) = \fracpartial{\Vapprox(\vecx,\Weights)}{\vecx}$, and this is what the VGL algorithms learn.  Since $\Vapprox(\vecx,\Weights)$ is defined to be smooth, the approximate value-gradient always exists.  

{\noindent \bf Greedy Policy.}  
We define a greedy policy $\pi(\vecx,\Weights)$ on the {\it approximate} value function $\Vapprox(\vecx, \Weights)$ by: 
\begin{equation}\pi(\vecx,\Weights)= \arg \max_{\veca \in \ActionSpace} (r(\vecx,\veca)+\df \Vapprox(f(\vecx,\veca), \Weights)) \label{eqn:greedyPolicy} \end{equation}
  
The greedy policy is a one-step look-ahead that decides which action to take, based only on the model and $\Vapprox$.  Since for a greedy policy, the actions are dependent on $\Vapprox(\vecx,\Weights)$ and state, it follows that $\pi = \pi(\vecx,\Weights)$.  This dependency on $\Weights$ distinguishes how we notate the greedy policy (i.e. $\pi(\vecx,\Weights)$) from a general (non-greedy) policy (i.e. $\pi(\vecx, \Weightsz)$).  We extend the definition of the value function $\Vpi(\vecx, \Weightsz)$ to also apply to the greedy policy,  and we write this as $\Vpi(\vecx, \Weights)$.

{\noindent \bf Greedy Trajectory.} A {\it greedy trajectory} is one that has been generated by the greedy policy.   Since the greedy policy depends upon the same weight vector as $\Vapprox(\vecx, \Weights)$, any modification to the weight vector $\Weights$ will immediately both change $\Vapprox(\vecx,\Weights)$ and move all greedy trajectories.  Hence we say $\Vapprox$ and the greedy policy are {\it tightly coupled};  it is the {\it same} weight vector $\Weights$ used in $\Vapprox(\vecx,\Weights)$ and the greedy policy $\pi(\vecx,\Weights)$.

{\noindent \bf Trajectory Shorthand Notation.}  For a given trajectory through states $(\vecx_0, \vecx_1, \ldots, \vecx_F)$ with actions $(\veca_0, \veca_1, \ldots, \veca_{F-1})$, and for any function defined on state space (e.g. including $\Vapprox(\vecx,\Weights)$, $\Gapprox(\vecx,\Weights)$, $\Rpi(\vecx,\Weights)$, $r(\vecx,\veca)$ and the functions defined later in the paper) we use a subscript of $t$ on the function to indicate that the function is being evaluated at $(\vecx_t, \veca_t, \Weights)$.  For example, $r_t=r(\vecx_t, \veca_t)$, $\Gapprox_t=\Gapprox(\vecx_t, \Weights)$ and $({\Rpi})_t$=${\Rpi}(\vecx_t,\Weights)$.  Note that this shorthand does not mean that these functions are functions of $t$.

{\noindent \bf Trajectory Shorthand Notation for Partial Derivatives.}  We use brackets with a subscripted $t$ to indicate that a partial derivative is to be evaluated at time step $t$ of a trajectory.  For example, $\fpevalat{\Gapprox}{\Weights}{t}$ is shorthand for $\left. \fracpartial{\Gapprox}{\Weights} \right|_{(\vecx_t,\Weights)}$, i.e. the function $\fracpartial{\Gapprox}{\Weights}$ evaluated at $(\vecx_t, \Weights)$.  Also, for example, $\fpevalat{f}{\veca}{t}=\left. \fracpartial{f}{\veca} \right|_{(\vecx_t, \veca_t)}$; 
and similarly for other partial derivatives including $\fpevalat{r}{\vecx}{t}$ and $\fpevalat{\Rpi}{\Weights}{t}$.

{\noindent \bf Matrix-vector notation.} Throughout this paper, a convention is used that all defined vector quantities are columns, whether they are coordinates, or derivatives with respect to coordinates.  Also any vector becomes transposed (becoming a row) if it appears in the numerator of a differential.  Upper indices indicate the component of a vector or matrix.  For example, $\vecx_t$ is a column; $\Weights$ is a column; $\Gapprox_t$ is a column; $\fpevalat{\Rpi}{\Weights}{t}$ is a column; $\fpevalat{f}{\vecx}{t}$ is a matrix with element $(i,j)$ equal to $\fpevalat{f(\vecx, \veca)^j}{\vecx^i}{t}$; $\fpevalat{\Gapprox}{\Weights}{t}$ is a matrix with element $(i,j)$ equal to $\fpevalat{\Gapprox^j}{\Weights^i}{t}$.  An example product is $\fpevalat{f}{\veca}{t}\Gapprox_{t+1}=\sum_i \fpevalat{f^i}{\veca}{t}\Gapprox^i_{t+1}$.  An example second derivative of a scalar is $\left(\fracpartial{^2\Vapprox}{\Weights \partial\vecx}\right)^{ij}=\left(\fracpartial{}{\Weights}\fracpartial{\Vapprox}{\vecx}\right)^{ij}=\fracpartial{}{\Weights^i}\fracpartial{\Vapprox}{\vecx^j}=\fracpartial{\Gapprox^j}{\Weights^i}$.

{\noindent \bf Approximate Q Value function.}  Since the quantity in the right-hand side of the greedy policy (eq. \ref{eqn:greedyPolicy}) comes up often, we define a function specifically for it.  The approximate Q Value function \citep{watkins92} is defined as 
\begin{equation} \Qapprox (\vecx,\veca,\Weights)=r(\vecx,\veca)+\df \Vapprox(f(\vecx,\veca), \Weights) \label{eqn:Qapprox} \end{equation}  
The greedy policy therefore maximises this quantity, i.e. the greedy policy is such that $\pi(\vecx,\Weights)= \arg \max_{\veca \in \ActionSpace} (\Qapprox (\vecx, \veca, \Weights))$.

We will often also need the derivative $\fracpartial{\Qapprox}{a}$ which is 
\begin{align}
 \fpevalat{\Qapprox}{\veca}{t}&=\fpevalat{r}{\veca}{t}+\df \fpevalat{f}{\veca}{t}\Gapprox_{t+1} \label{eqn:dQda}
\end{align}

\subsection{Target Values} \label{sec:targetValues}
These are useful concepts for understanding value-learning, which we need to define here because we will later differentiate them to make the targets for the VGL algorithm.

For a trajectory found by a greedy policy $\pi(\vecx,\Weights)$ on $\Vapprox(\vecx,\Weights)$, we define the ``target-value function'' $\Vtarget(\vecx, \Weights)$ recursively as 
\begin{subequations}
\begin{equation}\Vtarget(\vecx, \Weights)=r(\vecx, \pi(\vecx,\Weights))+\df \left(\lambda \Vtarget(f(\vecx,\pi(\vecx,\Weights)), \Weights)+(1-\lambda)\Vapprox(f(\vecx,\pi(\vecx,\Weights)), \Weights) \right)\label{eqn:lbq}\end{equation}	
with $\Vtarget(\vecx_F, \Weights)=0$ and where $\lambda \in [0,1]$ is a fixed global constant (described further below).  To calculate $\Vtarget$ for a particular point $\vecx_0$ in state space, it is necessary to run and cache a whole trajectory starting from $\vecx_0$ under the greedy policy $\pi(\vecx,\Weights)$, and then work backwards along it applying the above recursion; thus $\Vtarget(\vecx,\Weights)$ is defined for all points in state space.  

For a given trajectory, using shorthand notation the above equation simplifies to \begin{equation}                                                                                                                                                                                                                      
\Vtarget_t=r_t+\df \left( \lambda \Vtarget_{t+1}+(1-\lambda)\Vapprox_{t+1} \right) \label{eqn:lbqShorthand}
\end{equation}
\end{subequations}

We refer to the values $\Vtarget_t$ simply as the ``target values'' since the objective of VL is to make $\Vapprox_t$ equal to $\Vtarget_t$ for all $t$ and along all greedy trajectories, since then:
\begin{align}
 & \Vapprox_t= \Vtarget_t & \forall \vecx_0, \forall t & \nonumber \\
 \Longleftrightarrow \ \ & \Vapprox_t=r_t+\df \Vapprox_{t+1}  & \forall  \vecx_0, \forall t & \text{ by Eq. \ref{eqn:lbqShorthand}}  \nonumber \\
 \Longleftrightarrow \ \ & \Vapprox(\vecx,\Weights)=r(\vecx, \pi(\vecx,\Weights))+\df \Vapprox(f(\vecx, \pi(\vecx,\Weights)),\Weights) & \forall  \vecx  & \nonumber 
\\ \Longleftrightarrow \ \ & \Vapprox(\vecx,\Weights)=\Vpi(\vecx, \Weights)& \forall  \vecx  & \text{ by Eq. \ref{eqn:Rpi_definition}}\nonumber
\end{align}
So when coupled with the greedy policy (Eq. \ref{eqn:greedyPolicy}), $\Vapprox$ satisfies Bellman's Optimality Condition, and so the greedy policy will be an optimal policy.

We point out that since $\Vtarget$ is dependent on the actions and on $\Vapprox(\vecx,\Weights)$, it is not a simple matter to attain the objective $\Vapprox \equiv \Vtarget$, since changing $\Vapprox$ infinitesimally will immediately move the greedy trajectories (since they are tightly coupled), and therefore change $\Vtarget$;  these targets are moving ones.  So we should only try to move the values $\Vapprox$ {\it slowly} towards their targets.  For example a VL function approximator weight update to do this could be:
\begin{equation} \Delta \Weights = \alpha \sum_{t= 0}^{F-1} \fpevalat{\Vapprox}{\Weights}{t}(\Vtarget_t-\Vapprox_t) \label{eqn:tdWeightUpdate} \end{equation}
where $\alpha$ is a small positive constant.  

The choice of the constant $\lambda$ can be seen from Eq. \ref{eqn:lbqShorthand} to affect the targets and hence affect learning by Eq. \ref{eqn:tdWeightUpdate}.  If $\lambda=0$ then the recursion in Eq. \ref{eqn:lbqShorthand} is not needed, and the weight update will force the estimate $\Vapprox_t$ to move towards the estimate $\Vapprox_{t+1}$.  Updating an estimate based on an estimate like this is commonly called ``bootstrapping'', and we call $\lambda$ the bootstrapping parameter.  If $\lambda=1$ then the recursion in Eq. \ref{eqn:lbqShorthand} is fully used, giving $\Vtarget(\vecx, \Weights) \equiv \Vpi(\vecx, \Weights)$, and the estimate $\Vapprox_t$ will be updated to move towards the actual total reward received until terminating.  For other values of $0\leq\lambda\leq1$ we get a smooth blending between these two cases as can be seen by Eq. \ref{eqn:lbqShorthand}.

The function $\Vtarget$ is identical to the ``$\lambda$-Return'' \cite{Watkins1989}, as proven in \ref{sec:vDashEquivalenceToLambdaReturn}, but the $\Vtarget$ recursive formula is more succinct.  The weight update of Eq. \ref{eqn:tdWeightUpdate} is a succinct statement of the TD($\lambda$) algorithm in batch update mode, as also proven in \ref{sec:vDashEquivalenceToLambdaReturn}.  

The use of $\Vtarget$ greatly simplifies the analysis of value functions and value-gradients.  Having it defined in this recursive form allows us to easily differentiate it to form the target value-gradient, and hence define the VGL algorithms.  It would not be straightforward to define the VGL algorithms using the traditional formulation of the ``$\lambda$-Return'' described in \ref{sec:vDashEquivalenceToLambdaReturn}.

\subsection{Saturated Actions} \label{sec:saturatedActions}
An extra complication arises if actions are bounded, e.g. if the constraints $(\veca_t)^i \in [-1,1]$ are present for some action components.  These need handling carefully to be able to differentiate the policy function later in this paper.  For example if an action component represents the steering wheel of a car, then we say that action component is {\it saturated} when the steering wheel is rotated to its full limit in either direction, with pressure being applied against that limit.  Formally, for the greedy policy, when the constraints $(\veca_t)^i \in [-1, 1]$ are present for some action component $(\veca_t)^i$, we say the action component is {\it saturated} if $\left|(\veca_t)^i \right|=1$ and $\fpevalat{\Qapprox }{\veca^i}{t} \neq 0$.  If either of these conditions is not met, or the constraints are not present, then the action component $(\veca_t)^i$ is not saturated.  

We sometimes want to refer to just the unsaturated components of action vector $\veca$.  We use the notation $\ua$ to denote a vector of just the unsaturated components of $\veca$, i.e. this vector often has lower dimension than $\veca$ as any saturated components have been removed.  If there are no saturated components then $\ua=\veca$.

We note some useful Lemmas about saturated actions and the greedy policy.\footnote{These lemmas could be skipped on a first reading and just referred to as needed later in the paper.}

\begin{footnotesize}
\begin{lem}
\label{lem:lemmaGPzeroDeriv}If $(\veca_t)^i$ is saturated, then, whenever they exist,  $\fpevalat{\pi^i}{\vecx}{t} = 0$ and $\fpevalat{\pi^i}{\Weights}{t} = 0$.\end{lem}  This follows from the definition of saturated action components above:  Imagine if the steering wheel is fully turned to the right, with pressure applied.  Because of the pressure applied, any infinitesimal changes to the circumstances will not change the position of the steering wheel.

\noindent Note that $\fpevalat{\pi}{\vecx}{t}$ and $\fpevalat{\pi}{\Weights}{t}$ may not exist, for example, if there are multiple joint maxima in $\Qapprox (\vecx,\veca,\Weights)$ with respect to $\veca$.  Then an infinitesimal change to the $\Qapprox$ function could cause the maximum to flip discontinuously from one of these maxima to the other.

\begin{lem} \label{lem:lemmaSignOfdQda} If an action component $(\veca_t)^i$ chosen by a greedy policy is saturated, then $(\veca_t)^i=1 \Rightarrow \fpevalat{\Qapprox}{\veca^i}{t}>0$; and $(\veca_t)^i=-1 \Rightarrow \fpevalat{\Qapprox}{\veca^i}{t}<0$.\end{lem}  The first of these two implications has to be true since for a saturated action $\fpevalat{\Qapprox}{\veca^i}{t}\neq0$ by definition, and if $\fpevalat{\Qapprox}{\veca^i}{t}<0$ at $(\veca_t)^i=1$ then the maximum of $\Qapprox$ would not be at $(\veca_t)^i=1$, which contradicts the greedy policy.  The second implication is true for the same reason with the situation reversed.

\begin{lem} \label{lem:lemmaGPnegdef}If an action $\veca_t$ chosen by a greedy policy has some unsaturated components, then $\fpevalat{\Qapprox }{\ua}{t} = \vec{0}$ and $\ddQduaSqaredt$ is a negative semi-definite matrix.\end{lem}  The greedy policy has found an action somewhere in the middle of the vector space that contains $\ua$.  So this is an ordinary local maxima of a surface, hence possesses the claimed properties.

\begin{lem} \label{lem:lemmaDPidXdQdaZero} For any action $\veca_t$ chosen by the greedy policy, regardless of whether any components are saturated or not, whenever $\fpevalat{\pi}{\vecx}{t}$ exists, we have $\fpevalat{\pi}{\vecx}{t}\fpevalat{\Qapprox }{\veca}{t}=\vec{0}$.\end{lem}

\begin{pf}$\fpevalat{\pi}{\vecx}{t}\fpevalat{\Qapprox }{\veca}{t}=\sum_i \fpevalat{\pi^i}{\vecx}{t}\fpevalat{\Qapprox }{\veca^i}{t}$.  For each term of this sum, the greedy policy implies either $\fpevalat{\pi^i}{\vecx}{t}=\vec{0}$ 
(in the case that action component $(\veca_t)^i$ is saturated and
$\fpevalat{\pi^i}{\vecx}{t}$ exists, by Lemma \ref{lem:lemmaGPzeroDeriv}), or
$\fpevalat{\Qapprox }{\veca^i}{t}=0$ (in the case that $(\veca_t)^i$ is not saturated, by Lemma
\ref{lem:lemmaGPnegdef}). Hence each term of the sum is zero, hence the sum is zero.  \BlackBox \end{pf}
\end{footnotesize}

\subsection{Target Value-Gradient, ($\Gtarget_t$)} \label{sec:targetValueGradient}
We now define the target vectors $\Gtarget_t$ that will be used as the VGL objective which is to achieve $\Gapprox_t=\Gtarget_t$ for all $t>0$ along a greedy trajectory.  We first define it for $\lambda \in (0,1]$ and afterwards extend the definition to include $\lambda=0$. 

For $\lambda \in (0,1]$, we define the function $\Gtarget(\vecx,\Weights)=\fracpartial{\Vtarget(\vecx,\Weights)}{\vecx}$, which gives:
\begin{small}
\begin{align}\Gtarget_t 
=&\left(\fracpartial{}{\vecx}\left(r(\vecx, \pi(\vecx,\Weights))
+\df \lambda \Vtarget(f(\vecx,\pi(\vecx,\Weights)), \Weights)+(1-\lambda)\Vapprox(f(\vecx,\pi(\vecx,\Weights)), \Weights) \right)\right)_t & \text{(by Eq. \ref{eqn:lbq})}\nonumber \\
=&\left(\fpevalat{r}{\vecx}{t} +  \fpevalat{\pi}{\vecx}{t}\fpevalat{r}{\veca}{t}\right) 
&\nonumber\\&
+ \df \left(\fpevalat{f}{\vecx}{t} +  \fpevalat{\pi}{\vecx}{t}\fpevalat{f}{\veca}{t}\right)
	 \left(\lambda \Gtarget_{t+1}+(1-\lambda)\Gapprox_{t+1}\right) & \text{(by chain rule)}\label{eqn:vgbp}\end{align}
\end{small}
with $\Gtarget_{F}=\vec{0}$, and assuming all derivatives in this equation exist (these existence conditions are discussed further below).  

This recursive formula takes a known target value-gradient at the end point of a trajectory ($\Gtarget_F=\vec{0}$), and works it backwards along the trajectory rotating and incrementing it as appropriate, to give the target value-gradient at each time step.  

For $\lambda=0$, we modify the above definition slightly to make the definition independent of the existence of $\fpevalat{\pi}{\vecx}{t}$.  We first note that in the special case where $\lambda=0$, Eq. \ref{eqn:vgbp} simplifies as follows:
\begin{small}
\begin{align}
 \Gtarget_t &= \left(\fpevalat{r}{\vecx}{t} +  \fpevalat{\pi}{\vecx}{t}\fpevalat{r}{\veca}{t}\right)
	+ \df \left(\fpevalat{f}{\vecx}{t} +  \fpevalat{\pi}{\vecx}{t}\fpevalat{f}{\veca}{t}\right)
	 \Gapprox_{t+1} &\text{by Eq. \ref{eqn:vgbp}}\nonumber \\
  &=\fpevalat{r}{\vecx}{t} + \df \fpevalat{f}{\vecx}{t} \Gapprox_{t+1}
    +\fpevalat{\pi}{\vecx}{t}\left(\fpevalat{r}{\veca}{t}+\df \fpevalat{f}{\veca}{t}\Gapprox_{t+1}\right) & \nonumber \\
  &=\fpevalat{r}{\vecx}{t} + \df \fpevalat{f}{\vecx}{t} \Gapprox_{t+1}
    +\fpevalat{\pi}{\vecx}{t}\fpevalat{\Qapprox}{\veca}{t} &\text{by eq. \ref{eqn:dQda}} \nonumber \\
  &=\fpevalat{r}{\vecx}{t} + \df \fpevalat{f}{\vecx}{t} \Gapprox_{t+1}
    &\text{by Lemma \ref{lem:lemmaDPidXdQdaZero}} \label{eqn:vgbp_lambda0}
\end{align}
\end{small}

In this last line there was the assumption that $\fpevalat{\pi}{\vecx}{t}$ exists, but for $\lambda=0$ we simply define $\Gtarget_t$ to be equal to this last line.  Thus $\Gtarget_t$ is always defined and exists for $\lambda=0$.  This matches the target value-gradient that Werbos uses in the algorithms DHP and GDHP (Eq. 28 of \cite{werbos98adap-org}).  

In the special case of $\lambda=1$, $\Gtarget_t$ becomes identical to $\fpevalat{\Rpi}{\vecx}{t}$ (to see this, remember that for $\lambda=1$, we have $\Vtarget(\vecx, \Weights) \equiv \Vpi(\vecx,\Weights)$).

All terms of Eq. \ref{eqn:vgbp} are obtainable from knowledge of the model functions and the policy.  For obtaining the term $\fracpartial{\pi}{\vecx}$ it is usually preferable to have the greedy policy written in analytical form (e.g. the policy used by \cite{doya00reinforcement}).  Alternatively, using a derivation similar to that of Lemma \ref{lem:lemma2}, it can be shown that, when it exists, $\fpevalat{\pi}{\vecx}{t}$ is such that all saturated components satisfy 
$\fpevalat{\pi^i}{\vecx}{t}=\vec{0}$ and all unsaturated components satisfy:
\begin{equation} 
\fpevalat{u(\pi)}{\vecx}{t}=
-\df \fpevalat{^2\Qapprox }{\vecx \partial \ua}{t}\ddQduaSqaredt^{-1} \label{eqn:dpidx}
\end{equation}
if  $\ddQduaSqaredt^{-1}$ exists.

{\noindent \bf Existence conditions for the Target Value-Gradient}  
The target value-gradient is a key concept of the VGL method, so we should check under which conditions it exists. 

If $\lambda=0$ then $\Gtarget_t$ is defined to always exist by Eq. \ref{eqn:vgbp_lambda0}.  If $\lambda>0$ and if $\fracpartial{\pi}{\vecx}$ does not exist at some time step, $t_0$, of the trajectory, then $\Gtarget_t$ is not defined for all $t \leq t_0$.  Conditions in which $\fracpartial{\pi}{\vecx}$ might not exist are mentioned in Lemma \ref{lem:lemmaGPzeroDeriv} and Eq. \ref{eqn:dpidx}.

It may be that in some problems the environment and model functions make it so that $\Gtarget_t$ does not exist, even though the model functions are designed to be differentiable.  For example, if an agent is at the boundary between a terminal state and a non-terminal state, and its velocity is zero, then depending on which way it goes next will determine whether the trajectory terminates or not.  Hence the total reward could be vastly different in those two cases, and so the function $\Vpi$ is not differentiable with respect to $\vecx$ at that point.  These bifurcation points are hopefully rare in state space in most problems.  

The above rare occurrences of the non-existence of $\Gtarget$ do not affect the two main theoretical results of this paper (Sections \ref{sec:optimality} and \ref{sec:pgl}), since both results talk about consequences of when the target value gradient {\it does} exist.

However certain problems would not be suitable for the VGL method without their reformulation, for example if the total reward was defined to be equal to the {\it integer} number of time-steps in a trajectory.  This total reward function is a step function, so does not give any useful derivative for learning.  Even though time needs to be discretized for simulation purposes, a calculation of the actual continuous value of time would be needed to make a useful differentiable total reward function.  If the problem is modified so that the model functions more accurately reflect the underlying continuous time process then the VGL method will work.  As a rule of thumb, if a problem is suitable for PGL methods, then it will be suitable to work on VGL methods.

\subsection{The VGL($\lambda$) Learning Algorithm} \label{sec:theVGLalgorithm}
The objective for any VGL algorithm is to attain $\Gapprox_t =\Gtarget_t$ for all $t>0$ along a greedy trajectory.   It is proven in section \ref{sec:optimality} that this objective is sufficient to ensure the trajectory is locally extremal, and often locally optimal.  As with the objective for learning the targets $\Vtarget$, it should be noted that the VGL objective is not straightforward to achieve since the targets $\Gtarget_t$ are moving ones and are highly dependent on $\Weights$.  Hence we must use a weight update to {\it slowly} move the approximated gradients towards their targets:

\begin{equation} \Delta \Weights = \alpha \sum_{t=0}^{F-1} \fpevalat{\Gapprox}{\Weights}{t} \Omega_t (\Gtarget_t-\Gapprox_t) \label{eqn:NonResidGrads} \end{equation}

This equation defines the VGL($\lambda$) algorithm.  It is based on the GDHP and DHP algorithms \cite{hic92,werbos98adap-org}. Our definition of $\Gtarget$ in Eq. \ref{eqn:vgbp} extends Werbos' methods which were defined for only $\lambda=0$, to work with any $\lambda$.  Two implementations of this algorithm are given in \ref{sec:algorithmicImplementations}.

The $\Omega_t$ matrix is any positive definite matrix (as introduced by \cite{werbos98adap-org}), arbitrarily chosen by the experimenter.  $\Omega_t$ is included for generality, since the presence of {\it any} positive definite matrix here in the equation will force every component of $\Gapprox_t$ to move towards the corresponding component of $\Gtarget_t$ (in any basis).  For simplicity $\Omega_t$  is often just taken to be the identity matrix for all $t$ (as in Werbos' algorithm DHP).  One use for making $\Omega_t$ arbitrary could be for the experimenter to be able to compensate explicitly for any rescalings of the state space axes.  

It seems an inspired choice by Werbos to have included the matrix $\Omega_t$ at all, since in Section \ref{sec:pgl} it spontaneously appears in a PGL weight update, giving us an explicit formula we can use to specify $\Omega_t$ when $\lambda=1$.  In this case it suffices for $\Omega_t$ to be positive semi-definite.

Any VGL algorithm is going to involve using the matrices $\fpevalat{\Gapprox}{\Weights}{t}$ and/or $\fracpartial{\Gapprox}{\vecx}$ which, for neural-networks, involves second order back-propagation.  This is described in chapter 10 of \cite{hic92}.  

\section{The Local Optimality of the Value-Gradient Learning Objective} \label{sec:optimality}
\label{sec:optimalityProof}

In this section we define locally optimal trajectories and prove that if the VGL objective is achieved, i.e. if $\Gtarget_t=\Gapprox_t$ for all $t$ along a greedy trajectory, then that trajectory is locally extremal, and in certain situations, locally optimal.

\vskip 3 mm
\noindent

We first define the total reward for a given trajectory that is irrespective of the policy that was used to find it.  For any trajectory starting at state $\vecx_0$ and following actions $(\veca_0, \veca_1, \ldots, \veca_{F-1})$ until reaching a terminal state under the given model, the total reward encountered is given by the function: \begin{eqnarray}
R(\vecx_0, \veca_0, \veca_1, \ldots, \veca_{F-1}) &=&\sum_{t = 0}^{F-1} \dfa{^t} r(\vecx _{t}, \veca_{t})   \nonumber \\
 &=& r(\vecx_0, \veca_0)+\df R(f(\vecx_0,\veca_0), \veca_1, \veca_2, \ldots, \veca_{F-1}) \label{eqn:Raaa}  \end{eqnarray}
with $R({\vecx_F})=0$.  Thus $R$ is a function of the arbitrary starting state $\vecx_0$ and the actions.  We extend the trajectory shorthand notation to include the function $R$, so that for any given trajectory, $R_t\equiv R(x_t, \veca_t, \veca_{t+1},\ldots, \veca_{F-1})$.  This enables us to define the partial derivatives as $\fpevalat{R}{\vecx}{t}=\fracpartial{R_t}{\vecx_t}$ and $\fpevalat{R}{\veca}{t}=\fracpartial{R_t}{\veca_t}$.

\noindent{\bf Locally Optimal Trajectories.}  
We define a trajectory parameterised by values $(\vecx _0, \veca_0, \veca_1, \veca_2, \ldots)$ to be locally optimal if $R(\vecx_0, \veca_0, \veca_1, \veca_2, \ldots)$ is at a local maximum with respect to the parameters $(\veca_0, \veca_1, \veca_2, \ldots)$,
subject to the constraints (if present) that $(\veca_t)^i \in [-1, 1]$ for each action component $i$.

\noindent{\bf Locally Extremal Trajectories (LET).}  
We define a trajectory parameterised by values $(\vecx _0, \veca_0, \veca_1, \veca_2, \ldots)$ to be locally extremal if, for all $t$ and all action components $i$,
\begin{equation}\begin{cases}
  \fpevalat{R}{\veca^i}{t}=0 & \text{if \((\veca_t)^i\) is not saturated} \\
  \fpevalat{R}{\veca^i}{t}>0 & \text{if \((\veca_t)^i\) is saturated and \((\veca_t)^i=1\)} \\
  \fpevalat{R}{\veca^i}{t}<0 & \text{if \((\veca_t)^i\) is saturated and \((\veca_t)^i=-1\).} \\
  \end{cases}\label{eqn:letSufficientConditions}\end{equation}
In the case that all the actions are unbound, this criterion for a
LET simplifies to that of just requiring $\fpevalat{R}{\veca}{t}=\vec{0}$ for all $t$. 

Having the possibility of bound actions introduces the extra complication of saturated
actions.  The second condition in Eq. \ref{eqn:letSufficientConditions} 
 can be understood by the steering wheel analogy given before (in the definition of saturated action components); if the action component $(\veca_t)^i=1$ is saturated then the steering wheel is fully turned to the right {\it with pressure}, implying we would like to turn the car even more in that direction if that were possible (even though it isn't), which literally means that $\fpevalat{R}{\veca^i}{t}>0$.  The third condition in Eq. \ref{eqn:letSufficientConditions} is simply the reverse of this.

In fact, in this definition $R$ is locally optimal with respect to any saturated actions. 
Consequently, if all of the actions are fully saturated (a situation known as {\em bang-bang control}),  then this definition of a LET  provides a sufficient condition for a  locally optimal trajectory. 

\begin{lem} \label{lem:VGLobjectiveMetOverGreedyTraj}
For a greedy trajectory and any fixed $\lambda \in [0,1]$, if $\Gtarget_ t = \Gapprox_ t$ for all $t$ then $\Gtarget_t = \Gapprox_t = \fpevalat{R}{\vecx}{t}$ for all $t$. \end{lem}
\begin{pf}
This is proven by induction.  First we note that since $\Gapprox_t$ is defined to exist, then $\Gtarget_t$ must also exist (since $\Gtarget_ t = \Gapprox_ t$ for all $t$).  Next, for $\lambda \in (0,1]$, substituting $\Gtarget_t=\Gapprox_t$ into eq. \ref{eqn:vgbp} gives, \begin{align}                                                                                               
\Gapprox_t &= 
\fpevalat{\pi}{\vecx}{t}\left(\fpevalat{r}{\veca}{t}+\df \fpevalat{f}{\veca}{t}\Gapprox_{t+1}\right) + \fpevalat{r}{\vecx}{t}  +\df \fpevalat{f}{\vecx}{t} \Gapprox_{t+1}
&\nonumber\\
&=  \fpevalat{\pi}{\vecx}{t}\fpevalat{\Qapprox }{\veca}{t}  
+ \fpevalat{r}{\vecx}{t}  + \df \fpevalat{f}{\vecx}{t} \Gapprox_{t+1}
&\text{by eq. \ref{eqn:dQda}}\nonumber \\
 &=  \fpevalat{r}{\vecx}{t}  +\df \fpevalat{f}{\vecx}{t} \Gapprox_{t+1}
&\text{by Lemma \ref{lem:lemmaDPidXdQdaZero}}\nonumber
\end{align}
where in the application of Lemma \ref{lem:lemmaDPidXdQdaZero} we used the fact that $\fpevalat{\pi}{\vecx}{t}$ exists since $\Gtarget_t$ exists.  For $\lambda=0$, substituting $\Gtarget_t=\Gapprox_t$ into equation \ref{eqn:vgbp_lambda0} gives the same result again, i.e.
\begin{equation} 
\Gapprox_t = \fpevalat{r}{\vecx}{t} + \df \fpevalat{f}{\vecx}{t} \Gapprox_{t+1} \label{eqn:recurrenceRelationForG_onOptimalTrajectory}
\end{equation}  

So Eq. \ref{eqn:recurrenceRelationForG_onOptimalTrajectory} is valid for {\em all} $\lambda \in [0,1]$.

Also, differentiating Eq. \ref{eqn:Raaa} with respect to $\vecx$ gives \begin{equation} 
\fpevalat{R}{\vecx}{t}= \fpevalat{r}{\vecx}{t} + \df \fpevalat{f}{\vecx}{t}\fpevalat{R}{\vecx}{t+1} \nonumber
 \end{equation} 
\noindent So $\fpevalat{R}{\vecx}{t}$ and $\Gapprox_t$ have the same recursive definition.
Also their values at the final time step $t=F$ are the same, since
$\fpevalat{R}{\vecx}{F}= \Gapprox_F=\vec{0}$.  Therefore, by induction, $\Gtarget_t = \Gapprox_t = \fpevalat{R}{\vecx}{t}$ for all $t$.  \BlackBox  
\end{pf}

\begin{thm}
Any greedy trajectory satisfying $\Gtarget_ t =\Gapprox_ t$ (for all $t$) must be locally extremal. \label{thm:optimalityTheorem} 
\end{thm}
\begin{pf}
Since the greedy policy maximises $\Qapprox (\vecx_t, \veca_t, \Weights)$ with
respect to $\veca_t$ at each time-step $t$, we know at each $t$ and for each action component $i$,
\begin{equation}\begin{cases}
  \fpevalat{\Qapprox }{\veca^i}{t}=0 & \text{if \((\veca_t)^i\) is not saturated} \\
  \fpevalat{\Qapprox }{\veca^i}{t}>0 & \text{if \((\veca_t)^i\) is saturated and \((\veca_t)^i=1\)} \\
  \fpevalat{\Qapprox }{\veca^i}{t}<0 & \text{if \((\veca_t)^i\) is saturated and \((\veca_t)^i=-1\).} \\
  \end{cases} \label{eqn:gpConsequences} \end{equation}
These follow from Lemmas \ref{lem:lemmaGPnegdef} and \ref{lem:lemmaSignOfdQda}.  Therefore since,
\begin{align}
\fpevalat{R}{\veca}{t}&=\fpevalat{r}{\veca}{t}+\df \fpevalat{f}{\veca}{t}\fpevalat{R}{\vecx}{t+1} 
&\text{by differentiating eq. \ref{eqn:Raaa}} \nonumber \\
&=\fpevalat{r}{\veca}{t}+\df \fpevalat{f}{\veca}{t}\Gapprox_{t+1} &\text{by Lemma \ref{lem:VGLobjectiveMetOverGreedyTraj}}\nonumber \\
&=\fpevalat{\Qapprox }{\veca}{t} &\text{by eq. \ref{eqn:dQda}}\nonumber 
\end{align}
we have $\fpevalat{R}{\veca}{t} = \fpevalat{\Qapprox }{\veca}{t}$ for all $t$.  Therefore
the consequences of the greedy policy (Eq. \ref{eqn:gpConsequences}) become
equivalent to the sufficient conditions for a LET (Eq.
\ref{eqn:letSufficientConditions}), which implies the trajectory is a LET.
\BlackBox
\end{pf}

\begin{cor} 
If, in addition to the conditions of Theorem
\ref{thm:optimalityTheorem}, all of the
actions are saturated (bang-bang control), then the trajectory is locally
optimal.
\end{cor}
This follows from the definitions given above of a LET. \BlackBox

\begin{rmk}
According to the Bang-Bang Principle \cite{bangbangprinciple}, bang-bang control often arises in situations where the model functions are linear with respect to bound action vectors, or when the problem being solved is a time minimisation problem.  Hence it is often the case that all LETs found by this method are locally optimal.
\end{rmk}

\begin{rmk}
If the VGL objective (i.e. $\Gtarget_t=\Gapprox_t$ for all $t$) is achieved as the fixed point of a weight update that is gradient ascent on the total reward (e.g. such as the weight update of Eq. \ref{eqn:bpttgp} in Section \ref{sec:pgl}), then the LET must be locally optimal, because the objective was arrived at by gradient ascent on the total reward.
  
Since the weight update of Eq. \ref{eqn:bpttgp} is a special case of the VGL($\lambda$) weight update (Eq. \ref{eqn:NonResidGrads}), it is speculated, but still an open question, that any time the VGL objective is met by use of {\em any} instance of Eq. \ref{eqn:NonResidGrads} (i.e. while using any $\Omega_t$ matrix, or any $\lambda$), it could always produce locally {\em optimal} trajectories.
\end{rmk}

\begin{rmk}
We point out that the proof of Theorem \ref{thm:optimalityTheorem} is highly related to Pontryagin's maximum principle (PMP), since Eq. \ref{eqn:recurrenceRelationForG_onOptimalTrajectory} satisfies the PMP equation for a ``costate'' vector.  Therefore $\Gapprox_{t}$ is the costate vector of PMP, and the greedy policy (almost) forms the maximum condition of PMP (the only difference being that PMP is defined for continuous time systems).  This completes Pontryagin's conditions to be a LET.  However PMP still needs supplementing with Lemma \ref{lem:VGLobjectiveMetOverGreedyTraj} for it to be applicable for any $\lambda \in [0,1]$.  We did not use PMP explicitly because it only identifies LETs, and the way we have formulated the proof allows us to derive the corollary's extra conclusion for bang-bang control producing locally {\em optimal} trajectories.
\end{rmk}

\subsection{Discussion} \label{sec:optimalityDiscussion}

This local optimality proof is valid for any $\lambda \in [0,1]$.  This optimality condition only needs satisfying over a single trajectory, whereas for VL the corresponding optimality condition ($\Vtarget=\Vapprox$) needs satisfying over the whole of state space.  This implies that VGL methods have a much lesser requirement for exploration than VL methods do, since the {\em local} part of exploration comes for free with VGL methods.  What we mean by this is that provided the VGL learning algorithm makes progress towards achieving $\Gtarget_t=\Gapprox_t$ all along a greedy trajectory, the trajectory will make progress in bending itself towards a locally optimal shape, and this will happen without the need for any stochastic exploration.  We argue that this leads to greater efficiency for VGL compared to VL, and experiments confirm this in some simple problem domains, by several orders of magnitude in most cases \cite{fairbank08}.  

In comparison to VGL, the failure of VL without any exploration in a deterministic environment is dramatic and common, even when the value-function is {\em perfectly learned} along a single trajectory; examples are given in section 1.3 of \citep{fairbank08}. 

A separate efficiency issue is the algorithmic complexity of VL and VGL, and these are both the same ($O(\dim(\Weights))$ per time step) if Algorithm \ref{alg:VGLforwards} is used, but VGL is slower ($O(\dim(\Weights)(\dim(\vecx))^2)$ per time step) if the on-line implementation of Algorithm \ref{alg:VGLBackwards} is used.

The requirement of our optimality proof for episodic problems could be relaxed by introducing a discount factor $\gamma \in[0, 1]$ (see \cite{suttonbarto-1998} for details), and the proof can then be extended by altering the terminal step of the induction of Lemma \ref{lem:VGLobjectiveMetOverGreedyTraj} to instead use the boundary condition 
$\lim _{t \rightarrow \infty} \gamma ^t\Gtarget_t=\vec{0}$.  However it is not yet clear how to extend the proof to an {\it undiscounted} non-episodic problem.

The stochastic case for $\lambda=0$ is considered by \cite{werbos98adap-org}.

\section{The Relationship of VGL to Policy-Gradient Learning}\label{sec:pgl}
We now prove that the VGL($\lambda$) weight update of Eq. \ref{eqn:NonResidGrads}, with $\lambda=1$ and a carefully chosen $\Omega_t$ matrix, is equivalent to PGL on a greedy policy.  

PGL, sometimes also known as ``direct'' reinforcement learning,  is defined to be gradient ascent on $\Rpi(\vecx_0,\Weightsz)$ with respect to the weight vector $\Weightsz$ of a general policy, i.e. $\Delta \Weightsz=\alpha \fpevalat{\Rpi}{\Weightsz}{0}$ for some small positive constant $\alpha$.  PGL methods will naturally find local maxima of $\Rpi(\vecx_0,\Weightsz)$ and have good convergence properties. 

A very direct and efficient method to calculate the policy gradient, $\fpevalat{\Rpi}{\Weightsz}{0}$, when the model functions are known is backpropagation through time (BPTT) \cite{backproptime90} which we will follow here, and it is well suited to deterministic systems.  However most studies of PGL in the RL literature \citep{williams,Peters06policygradient} are designed for stochastic environments and unknown model functions, and they form the mean $\fpevalat{\left< \Rpi \right>}{\Weightsz}{0}$ after sampling many trajectories.  \cite{munos06} describes a rapid model learning method that finds the policy gradient after just one trajectory.

The required PGL gradient can be expanded as follows:

\begin{align}
\fpevalat{\Rpi}{\Weightsz}{t}=&  \left({ \fracpartial{}{\Weightsz} (r(\vecx,\pi(\vecx,\Weightsz))+\df \Rpi(f(\vecx,\pi(\vecx,\Weightsz)),\Weightsz))}\right)_t\nonumber & \text{(by Eq. \ref{eqn:Rpi_definition})}\\
=& \fpevalat{\pi}{\Weightsz}{t} \left( \fpevalat{r}{\veca}{t} +\df \fpevalat{f}{\veca}{t}\fpevalat{\Rpi}{\vecx}{t+1} \right)+\df \fpevalat{\Rpi}{\Weightsz}{t+1} &\text{(by chain rule)}\nonumber \end{align}

Expanding this recursion and substituting it into the gradient ascent equation gives,
\begin{align}
\Delta \Weightsz=& \alpha \sum _{t \ge 0} \dfa{^t}
\fpevalat{\pi}{\Weightsz}{t} \left( \fpevalat{r}{\veca}{t}
+\df \fpevalat{f}{\veca}{t}\fpevalat{\Rpi}{\vecx}{t+1} \right) \nonumber \end{align}

BPTT is merely an efficient implementation of this formula, often for cases where the policy $\pi(\vecx, \Weightsz)$ is provided by a neural-network \citep{backproptime90}, but also defined for a general differentiable policy.  We note that although this equation looks quite different from the more regularly used PGL equation of \cite{williams}, the two are mathematically equivalent since it is proven in \cite{williams} that their weight update is equal to $\fracpartial{\left<\Rpi \right>}{\Weightsz}$  

The above weight update was derived for a general policy.  We now switch to specifically consider PGL applied to a greedy policy, so that all instances of the weight vector $\Weightsz$ will be now replaced by $\Weights$:  
\begin{align}
\Delta \Weights=& \alpha \sum _{t \ge 0} \dfa{^t}
\fpevalat{\pi}{\Weights}{t} \left( \fpevalat{r}{\veca}{t}
+\df \fpevalat{f}{\veca}{t}\fpevalat{\Rpi}{\vecx}{t+1} \right) \label{eqn:bpttgp1} \end{align}
The equivalence we will show only holds when the actions are  unbound.  If bound actions are required then they could be implemented indirectly by applying an exponential cost function to the model function $r(\vecx, \veca)$ to penalise components of $\veca$ that get close to their desired limits, prohibiting the greedy policy from choosing actions beyond this range. 

Initially we only consider the case where $\fpevalat{\Rpi}{\Weights}{0}$ exists for a greedy trajectory, and also where hence $\fpevalat{\pi}{\Weights}{t}$ exists for all $t$.  The  terms $\fpevalat{\pi}{\Weights}{t}$ and $\fpevalat{r}{\veca}{t}$ can be reinterpreted under the greedy policy:

\begin{lem}
\label{lem:lemma1} The greedy policy implies, for an unbound action, 
$\fpevalat{r}{\veca}{t} =- \df \fpevalat{f}{\veca}{t}\Gapprox_{t+1}$.
\end{lem}
\begin{pf}
Substitute $\fpevalat{\Qapprox }{\veca}{t}=\vec{0}$ (Lemma \ref{lem:lemmaGPnegdef} for an unbound action) into Eq. \ref{eqn:dQda}. \BlackBox
\end{pf}

\begin{lem}
\label{lem:lemma2}
When $\fpevalat{\pi}{\Weights}{t}$ and $\ddQdaSqaredt^{-1}$ exist for an unbound action $\veca_t$, the greedy policy implies 
\begin{equation}
\fpevalat{\pi}{\Weights}{t}= -\df \fpevalat{\Gapprox}{\Weights}{t+1} \fpevalat{f}{\veca}{t}^T
\ddQdaSqaredt^{-1} \nonumber
\end{equation}
\end{lem}

\begin{pf}
We use implicit differentiation.  The dependency of $\veca_t=\pi(\vecx_t,\Weights)$ on $\Weights$ must be such that Lemma \ref{lem:lemmaGPnegdef} is always satisfied, i.e. so that $\fpevalat{\Qapprox }{\veca}{t} \equiv \vec{0}$, both before and after any infinitesimal change to $\Weights$. Therefore the function $\pi(\vecx_t,\Weights)$ must be such that,
\begin{align}
\vec{0} &= \fracpartial{}{\Weights} \left( \frac{\partial \Qapprox (\vecx_t,\pi(\vecx_t,\Weights),\Weights)
}{\partial \veca_t} \right) \nonumber \\
&= \fracpartial{}{\Weights} \left(\frac{\partial \Qapprox (\vecx_t,\veca_t,\Weights)}{\partial \veca_t} \right) + \fpevalat{\pi}{\Weights}{t}\frac{\partial}{\partial \veca_t} \left( \frac{\partial \Qapprox (\vecx_t,\veca_t,\Weights)}{\partial \veca_t} \right) & \text{(by chain rule)}\nonumber &\\
&= \fracpartial{}{\Weights} \left( \fpevalat{r}{\veca}{t} + \df \fpevalat{f}{\veca}{t} \Gapprox_{t+1} \right) + \fpevalat{\pi}{\Weights}{t}\ddQdaSqaredt& \text{(by Eq. \ref{eqn:dQda})}\nonumber \\
&=\fracpartial{}{\Weights} \left( \fpevalat{r}{\veca}{t} +  \df \sum_i \fpevalat{(f)^i}{\veca}{t} (\Gapprox_{t+1})^i \right) + \fpevalat{\pi}{\Weights}{t}\ddQdaSqaredt&\text{(inner product)}\nonumber \\
&= \df   \sum_i \fpevalat{(f)^i}{\veca}{t} \fracpartial{(\Gapprox_{t+1})^i}{\Weights}  + \fpevalat{\pi}{\Weights}{t}\ddQdaSqaredt& \nonumber \\
&= \df \fpevalat{\Gapprox}{\Weights}{t+1} \fpevalat{f}{\veca}{t}^T + \fpevalat{\pi}{\Weights}{t}\ddQdaSqaredt &\text{(inner product)}\nonumber \end{align}
The penultimate line used the fact that $\fracpartial{r}{\veca}$ and $\fracpartial{f}{\veca}$ are not functions of $\Weights$.  Then solving the final line for $\fpevalat{\pi}{\Weights}{t}$ proves the lemma. \BlackBox 
\end{pf}

Substituting these Lemmas \ref{lem:lemma1} and \ref{lem:lemma2}, and $\fpevalat{\Rpi}{\vecx}{t}=\Gtarget_t$ with $\lambda =1$ (see Eq. \ref{eqn:vgbp}), into Eq. \ref{eqn:bpttgp1} gives:
\begin{eqnarray} 
\Delta \Weights&=& \alpha  \sum _{t \ge 0} \dfa{^{t}} \left( 
-\dfa{^2}\fpevalat{\Gapprox}{\Weights}{t+1} \fpevalat{f}{\veca}{t}^T
\ddQdaSqaredt^{-1}
\fpevalat{f}{\veca}{t} (-\Gapprox_{t+1}+\Gtarget_{t+1}) \right) \nonumber \\
& = & \alpha \sum _{t \ge 0} \dfa{^{t+2}} 
\fpevalat{\Gapprox}{\Weights}{t+1} \Omega _t(\Gtarget_{t+1}-\Gapprox_{t+1}) \label{eqn:bpttgp} \end{eqnarray}
\noindent where \begin{equation}
\Omega _t=-\fpevalat{f}{\veca}{t}^T
\ddQdaSqaredt^{-1}
\fpevalat{f}{\veca}{t} \label{eqn:Omega}, \end{equation}
and is positive semi-definite, by the greedy policy (Lemma \ref{lem:lemmaGPnegdef}).

Equation \ref{eqn:bpttgp} is identical to a VGL weight update equation (Eq. \ref{eqn:NonResidGrads}), with a carefully chosen matrix for $\Omega_t$, and $\lambda=1$, provided $\fpevalat{\pi}{\Weights}{t}$ and $\ddQdaSqaredt^{-1}$ exist for all $t$.  If $\fpevalat{\pi}{\Weights}{t}$ does not exist, then $\fracpartial{\Rpi}{\Weights}$ is not defined either.

This completes the demonstration of the equivalence of a value-function learning algorithm (VGL($1$), with the conditions stated above) to PGL (with greedy policy; when $\fracpartial{\Rpi}{\Weights}$ exists).

\subsection{Discussion} \label{sec:pglEquivalenceDiscussion}

If the RL problem is such that $\fpevalat{\pi}{\Weights}{t}$ always exists, then the good convergence guarantees of PGL will apply to VGL(1), under the above conditions. For certain simple problems this is always true, but significantly it is always true in the continuous time setting for VGL \citep{fairbank08}, where the value-gradient policy defined by \cite{doya00reinforcement} is used.  Conveniently, this policy also ensures the actions are always completely unsaturated, which was a condition for the PGL equivalence.

This equivalence result was surprising to the authors because it was thought that the VL and VGL weight updates (equations  \ref{eqn:tdWeightUpdate} and \ref{eqn:NonResidGrads}) were based on {\it gradient descent} on the error functions $\sum_t (\Vtarget_t-\Vapprox_t)^2$ and $\sum_t (\Gtarget_t-\Gapprox_t)^T \Omega_t (\Gtarget_t-\Gapprox_t)$, respectively.   In fact the weight update of Eq. \ref{eqn:tdWeightUpdate} is sometimes generally described as a gradient descent weight update of that error function (e.g. see chapter 8.2 of \cite{suttonbarto-1998}).  But neither weight update is true gradient descent, unless both $\lambda=1$ and the policy is fixed, i.e. non-greedy.  For example, equations  \ref{eqn:tdWeightUpdate} and \ref{eqn:NonResidGrads} have far fewer terms in them than would be found by differentiating the two error functions fully using the chain rule (e.g. \citep{iii95residual} describes the missing terms for a fixed policy with VL and $\lambda=0$; even more terms are missing if a tightly-coupled greedy policy is used, even when $\lambda=1$).  Therefore learning progress as measured by either error function is far from monotonic.  So this PGL-VGL equivalence proof is surprising because it shows these anomalies for VGL(1) are because the weight update is actually {\it gradient ascent} on $\Vpi$ when $\Omega_t$ is chosen carefully, and it neatly solves the monotonic progress problem for a tightly-coupled greedy policy.

It was also surprising to learn that PGL and value-function weight updates are not as fundamentally different to each other as we first thought.  It was not known that any PGL weight update, when applied to a greedy policy on an approximate value function, would be doing the same thing as any value function weight update; even if both had $\lambda=1$.  Of course they are usually {\it not} the same, unless this particular choice of $\Omega_t$ is chosen.  
The equivalence now creates a difficulty in distinguishing between PGL and VGL($1$) with this particular $\Omega_t$.  However, we describe the above weight update as a VGL update; it is of the same form as Eq. \ref{eqn:NonResidGrads} which was defined by \cite{werbos98adap-org}, i.e. prior to this equivalence being realised.

If $\lambda=1$ is used then Eq. \ref{eqn:Omega} is a good choice for $\Omega_t$, since it will ensure monotonic progress with respect to $\Rpi$, under the above conditions.  However Eq. \ref{eqn:Omega} means $\Omega_t$ can sometimes be very small, which could slow learning down.  Alternative choices for $\Omega_t$ (such as the identity matrix) may hence produce an aggressive speed up of learning, but will do so at the expense of monotonic progress.  
A less aggressive speed up method for the gradient ascent, such as conjugate gradients, could be used instead of using the identity matrix for $\Omega_t$.

This analysis has been successful for a tightly-coupled greedy policy and value function, which is unusual since most RL analyses of value-function updates in the literature so far have only been applicable for a ``fixed'' policy.  Interestingly, using a tightly-coupled value-function and greedy policy was necessary for the equivalence to hold.  

\cite{suttonMcallester} provides a related convergence result that also applies to the problem of concurrently updating $\Vapprox$ and $\pi$. Their result applies to an actor-critic architecture, and since this does not use a greedy policy, they avoid the need for considering $\fracpartial{\pi}{\Weights}$ through Lemma \ref{lem:lemma2}.  While our result compared to theirs is disadvantaged in that it is only valid for $\lambda=1$, some possible advantages of our method over theirs are as follows:  Their result is thought to apply only when the function approximator for $\Vapprox$ is linear in the same features of the state vector that the function approximator for the policy uses as input (see footnote 1 of \citep{suttonMcallester}).  Also the policy iteration algorithm that updates both function approximators requires that the function approximator for $\Vapprox$ is trained {\em to completion, over all of state space}, every time the policy changes, and this requirement is nested inside of the loop that updates the policy; so the whole thing is quite computationally expensive.  Finally, in order for the training process for $\Vapprox$ to have guaranteed convergence, it would have to be linear in $\Weights$ if $\lambda<1$ \citep{tsitsiklis96analysis}.

\section{Conclusions and Further Work} \label{sec:conclusions}

We have proven the local optimality of learning the target value gradients along a greedy trajectory (for any $\lambda$), argued the efficiency benefits of that approach, and have demonstrated the equivalence of VGL(1) to PGL.  In this research we have been interested in genuine theoretical challenges to understanding value-function learning with a greedy policy, regardless of whether by VL or VGL; particularly about how the greedy policy is affected by a weight update to $\Vapprox(\vecx, \Weights)$ (as derived in Lemma \ref{lem:lemma2}), and particularly about what exactly is required for an optimal trajectory (Section \ref{sec:optimalityProof}).

In further work we would like to extend the optimality proof of Section \ref{sec:optimality} to undiscounted non-episodic problems, and of course somehow work how to extend the convergence proof for $\lambda=1$ to include $\lambda<1$, which is unlikely with the weight update in its current form since divergence examples for this exist (e.g. section 4.3 of \cite{fairbank08}).

\section*{Acknowledgements}
We are very grateful to Peter Dayan, Paul Werbos, Csaba Szepesvari, R\'{e}mi Coulom, Lucian Busoniu and the reviewers for their discussions, suggestions and pointers for research on this topic.

\appendix

\section {Equivalence of $\Vtarget$ notation to the $\lambda$-Return} \label{sec:vDashEquivalenceToLambdaReturn}
Although it was first discovered by Watkins \citep{Watkins1989}, we use the definition of the $\lambda$-Return given by \cite{suttonbarto-1998}:
$ R_t^\lambda=(1-\lambda)\sum_{n=1}^\infty \lambda^{n-1}R_t^{(n)}$ with $R_t^{(n)}=\sum_{k=0}^{n-1} r_{t+k}+\Vapprox_{t+n}$.  We aim to show that $\Vtarget_t$ is identical to $R_t^\lambda$.  Expanding the definition of $R_t^\lambda$ gives
\begin{align}
 R_t^\lambda&=(1-\lambda)\sum_{n=1}^\infty \lambda^{n-1}\left(\sum_{k=0}^{n-1} r_{t+k}+\Vapprox_{t+n}\right) \nonumber \\
 &=(1-\lambda)\left(\lambda^0r_{t}+\lambda^1(r_{t}+r_{t+1})+\lambda^2(r_{t}+r_{t+1}+r_{t+2})+\ldots\right)
+(1-\lambda)\sum_{n=1}^\infty \lambda^{n-1}\Vapprox_{t+n}  \nonumber \\
 &=(1-\lambda)\sum_{n=0}^\infty \left(r_{t+n}\sum_{k=n}^\infty \lambda^k \right)
+(1-\lambda)\sum_{n=0}^\infty \lambda^{n}\Vapprox_{t+n+1} \nonumber  \\
 &=\sum_{n=0}^\infty \left(\lambda^{n} r_{t+n}  \right)+(1-\lambda)\sum_{n=0}^\infty \lambda^{n}\Vapprox_{t+n+1} \nonumber
\end{align}
Expanding the definition of $\Vtarget$ (Eq. \ref{eqn:lbqShorthand}) gives 
\begin{align}
\Vtarget_t&=r_t+\df \left( \lambda \Vtarget_{t+1}+(1-\lambda)\Vapprox_{t+1} \right) \nonumber \\
&=\sum_{n=0}^\infty \lambda^{n} \left(r_{t+n}+\df (1-\lambda)\Vapprox_{t+n+1} \right) \nonumber 
\end{align}

Thus $\Vtarget$ is identical to $R^\lambda$.  However since it uses a recursive notation, $\Vtarget$ is easier to differentiate than $R^\lambda$, enabling us to define $\Gtarget$.  The $\lambda$-Return provides an equivalent formulation for the algorithm TD($\lambda$) known as the ``forwards view of TD($\lambda$)'' \citep{suttonbarto-1998}.  This proves that Eq. \ref{eqn:tdWeightUpdate} is equivalent to the TD($\lambda$) weight update.

\section{A batch mode implementation, and an on-line implementation of the VGL($\lambda$) algorithm}
\label{sec:algorithmicImplementations}
We give two implementations of the VGL($\lambda$) algorithm which produce an identical weight update. Algorithm \ref{alg:VGLforwards} is the faster of the two, but requires storage of a whole trajectory and is batch-mode only.  Algorithm \ref{alg:VGLBackwards} can be used on-line and is more memory efficient since it does not store a whole trajectory, but is slower since it requires the manipulation of an ``eligibility trace'' matrix.

Here we use a shorthand notation as $\bigD{}{\vecx}\equiv \fracpartial{}{\vecx}+\fracpartial{\pi}{\vecx}\fracpartial{}{\veca}$.  We let $\dimWeights$ and $\dimX$ be the dimensions of $\Weights$ and $\vecx$ respectively.  $\gamma$ is a constant {\it discount factor}, where $\gamma \in [0,1]$.

Algorithm \ref{alg:VGLforwards} makes a direct implementation of Eq. \ref{eqn:NonResidGrads} by making a foward pass through the trajectory, storing all states and actions, followed by a backward pass through the trajectory accumulating $\Gtarget_t$ by the recursion in Eq. \ref{eqn:vgbp}. In this implementation, the second order derivatives of the approximate value function (i.e. $\fracpartial{\Gapprox}{\Weights}$ and $\fracpartial{\pi}{\vecx}$) are only required as an inner product with a vector.  This means that if a neural network is used for the function approximator, and if we use methods analogous to those used by \cite{pearlmutter94fast} or ch. 10 of \citep{hic92}, then these matrix-vector products can be evaluated in O($\dimWeights$) operations.  This means it takes O($F\dimWeights$) operations for the algorithm to run on a whole trajectory.

\begin{algorithm}[h]
\begin{algorithmic}[1]
\caption {VGL($\lambda$).  Batch-mode implementation.}
\label{alg:VGLforwards}
\begin{minipage}[b]{0.5\linewidth}
\STATE $t\leftarrow 0$, $\Delta \Weights\leftarrow \vec{0}$
\STATE \COMMENT {Unroll trajectory...}
\WHILE {not terminated($\vecx_t$)}
\STATE $\veca_t \leftarrow \pi(\vecx_t, \Weights)$
\STATE $\vecx_{t+1} \leftarrow f(\vecx_t, \veca_t)$
\STATE $t \leftarrow t+1$
\ENDWHILE
\STATE $F \leftarrow t$
\STATE $\vec{p} \leftarrow \vec{0}$
\end{minipage}
\hspace{0.5cm}
\begin{minipage}[b]{0.5\linewidth}
\STATE \COMMENT {Backwards pass...}
\FOR {$t=F-1$ to $0$ step $-1$}
\STATE $\Gtarget_t \leftarrow \bigDevalat{r}{\vecx}{t}+\gamma \bigDevalat{f}{\vecx}{t}\vec{p}$
\STATE $\Delta \Weights \leftarrow \Delta \Weights+\fpevalat{\Gapprox}{\Weights}{t} \Omega_t\left(\Gtarget_t-\Gapprox_t\right)$
\STATE $\vec{p}\leftarrow \lambda \Gtarget_t +(1-\lambda)\Gapprox_t$
\ENDFOR
\STATE $\Weights \leftarrow \Weights+\alpha \Delta \Weights$
\end{minipage}
\end{algorithmic} 
\end{algorithm}

Algorithm \ref{alg:VGLBackwards} accumulates all the weight updates in a single forward pass of the trajectory.  It requires no storage of the trajectory, so is more memory efficient, but requires more time to carry out matrix multiplications.  This algorithm requires the full $\fracpartial{\Gapprox}{\Weights}$ matrix which, for a neural network, would take O($\dimX\dimWeights$) operations to evaluate.  Hence the slowest steps in the algorithm would be the matrix-matrix multiplications of lines \ref{line:VGLBackwards:matrixmatrixmult1} and \ref{line:VGLBackwards:matrixmatrixmult2}, each taking O($(\dimX)^2\dimWeights$) operations.  Hence the total time for the algorithm to process a full trajectory is O($F(\dimX)^2\dimWeights$) operations.  

To derive this algorithm, we had to first rewrite Eq. \ref{eqn:vgbp} as follows:
\begin{align}
 \Gtarget_t=&\bigDevalat{r}{\vecx}{t}+ \gamma \bigDevalat{f}{\vecx}{t}\left(\lambda \Gtarget_{t+1}+(1-\lambda)\Gapprox_{t+1}\right) \nonumber \\
=&\bigDevalat{r}{\vecx}{t}+ \gamma \bigDevalat{f}{\vecx}{t}\Gapprox_{t+1}+\lambda \gamma  \bigDevalat{f}{\vecx}{t}\left( \Gtarget_{t+1}-\Gapprox_{t+1}\right) \nonumber \\
\Rightarrow \Gtarget_t-\Gapprox_t=&\left(\bigDevalat{r}{\vecx}{t}+ \gamma \bigDevalat{f}{\vecx}{t}\Gapprox_{t+1}-\Gapprox_t\right)+\lambda \gamma  \bigDevalat{f}{\vecx}{t}\left( \Gtarget_{t+1}-\Gapprox_{t+1}\right) \nonumber\\
=&\vecDelta_t+\lambda  \gamma \bigDevalat{f}{\vecx}{t}\left( \Gtarget_{t+1}-\Gapprox_{t+1}\right) \label{eqn:GdashMinusGrecursion}
\end{align}
where we define
\begin{align}
\vecDelta_t=\bigDevalat{r}{\vecx}{t}+ \gamma \bigDevalat{f}{\vecx}{t}\Gapprox_{t+1}-\Gapprox_t \label{eqn:VGL_delta}
\end{align} 
Unrolling the recursion in $(\Gtarget_t-\Gapprox_t)$ of Eq. \ref{eqn:GdashMinusGrecursion} gives
\begin{align}
\Gtarget_t-\Gapprox_t=\vecDelta_t+\lambda  \gamma \bigDevalat{f}{\vecx}{t}\vecDelta_{t+1}+\lambda^2  \gamma{^2}  \bigDevalat{f}{\vecx}{t}\bigDevalat{f}{\vecx}{t+1}\vecDelta_{t+2}+\ldots \nonumber
\end{align}
Then substituting this into the VGL($\lambda$) weight update equation (Eq. \ref{eqn:NonResidGrads}) and reordering the terms gives:
\begin{align}
\Delta \Weights &= \alpha \sum_{t=0}^{F-1} \fpevalat{\Gapprox}{\Weights}{t} \Omega_t \left(\vecDelta_t+\lambda  \gamma \bigDevalat{f}{\vecx}{t}\vecDelta_{t+1}+\lambda^2  \gamma{^2} \bigDevalat{f}{\vecx}{t}\bigDevalat{f}{\vecx}{t+1}\vecDelta_{t+2}+\ldots \right) \nonumber \\ 
& = \alpha \sum_{t=0}^{F-1} \left( E_t \vecDelta_t\right) \label{eqn:vglWeightUpdateWithEligibilityTrace} 
\end{align}
where $E_t$ is a matrix defined to be
\begin{align}E_t=& \fpevalat{\Gapprox}{\Weights}{t} \Omega_t
+\lambda  \gamma \fpevalat{\Gapprox}{\Weights}{t-1} \Omega_{t-1} \bigDevalat{f}{\vecx}{t-1}
+\lambda^2  \gamma{^2} \fpevalat{\Gapprox}{\Weights}{t-2} \Omega_{t-2} \left(\bigDevalat{f}{\vecx}{t-2}\bigDevalat{f}{\vecx}{t-1}\right)
 \nonumber \\ &
+\ldots+\lambda^t  \gamma{^t} \fpevalat{\Gapprox}{\Weights}{0} \Omega_{0} 
\left(\bigDevalat{f}{\vecx}{0}\bigDevalat{f}{\vecx}{1}\ldots \bigDevalat{f}{\vecx}{t-2}\bigDevalat{f}{\vecx}{t-1}\right)\nonumber
\end{align}
We see that $E_t$ can be defined more simply by a recursion:
\begin{equation}
E_t= \fpevalat{\Gapprox}{\Weights}{t} \Omega_t +\lambda  \gamma E_{t-1} \bigDevalat{f}{\vecx}{t-1} \label{eqn:vglEligibilityTrace}
\end{equation}
with $E_{-1}=0$.  We call the matrix $E_t$ an ``eligibility trace'' matrix because it acts similarly to the eligibility trace described for TD($\lambda$) \citep{sutton88learning}.  Algorithm \ref{alg:VGLBackwards} is then easily derived from Equations \ref{eqn:VGL_delta}, \ref{eqn:vglWeightUpdateWithEligibilityTrace} and \ref{eqn:vglEligibilityTrace}.

\begin{algorithm}[h]
\begin{algorithmic}[1]
\caption {VGL($\lambda$).  On-line implementation.}
\label{alg:VGLBackwards}
\begin{minipage}[b]{0.5\linewidth}
\STATE $E\leftarrow 0$ 
\COMMENT {$E \in \Re^{\dimWeights \times \dimX}$ is an ``eligibility trace'' workspace matrix.}
\STATE $t\leftarrow 0$, $\Delta \Weights\leftarrow \vec{0}$
\STATE {\bf while} not terminated($\vecx_t$) {\bf do}
\STATE $\ \ \ \veca_t \leftarrow \pi(\vecx_t, \Weights)$
\STATE $\ \ \ \vecx_{t+1} \leftarrow f(\vecx_t, \veca_t)$
\STATE $\ \ \ \vecDelta \leftarrow \bigDevalat{r}{\vecx}{t}-\Gapprox_t$
\STATE  \ \ \ {\bf if} not terminated($\vecx_{t+1}$) {\bf then}
\STATE \ \ \ \ \ \ $\vecDelta \leftarrow \vecDelta+ \gamma \bigDevalat{f}{\vecx}{t}\Gapprox_{t+1}$
\end{minipage}
\hspace{0.5cm}
\begin{minipage}[b]{0.5\linewidth}
\STATE \ \ \ {\bf end if}
\STATE $\ \ \ E\leftarrow  E+\fpevalat{\Gapprox}{\Weights}{t} \Omega_t$ \label{line:VGLBackwards:matrixmatrixmult1}
\STATE $\ \ \ \Delta \Weights \leftarrow \Delta \Weights+ E \vecDelta$ \label{line:VGLBackwards:matrixvectormult1}
\STATE $\ \ \ E\leftarrow \lambda  \gamma E \bigDevalat{f}{\vecx}{t}$ \label{line:VGLBackwards:matrixmatrixmult2}
\STATE $\ \ \ t \leftarrow t+1$
\STATE {\bf end while}
\STATE $\Weights \leftarrow \Weights+\alpha \Delta \Weights$, $\Delta \Weights\leftarrow \vec{0}$ 
\COMMENT {This line can be moved up one position if true on-line weight updating is required.}
\end{minipage}
\end{algorithmic} 
\end{algorithm}

Neither of the two implementations in this section attempts to learn the value gradient at the final time-step of a trajectory since it is prior knowledge that the target value gradient is always zero at any terminal state.  Hence we assume the function approximator for $\Vapprox(\vecx, \Weights)$ has been designed to explicitly return zero for all terminal states $\vecx$.


\bibliographystyle{elsarticle-num}
\bibliography{RLValueGradients}







\end{document}